\documentclass[journal,onecolumn]{IEEEtran}
\usepackage{authblk}
\usepackage{cite}
\usepackage{graphicx}
\usepackage{hyperref}
\usepackage{float}
\usepackage{soul}
\usepackage{xcolor}
\usepackage{array}
\usepackage{caption}
\usepackage{verbatim}
\usepackage{titlesec}
\usepackage{amssymb}
\usepackage{enumitem}
\usepackage{multirow}
\usepackage{tabularx}
\usepackage{rotating}
\usepackage{amsmath}
\usepackage{amssymb}
\usepackage{forest}
\usepackage{adjustbox}
\usepackage{makecell}

\usepackage{tikz}
\usetikzlibrary{shapes.geometric, arrows}

\tikzstyle{startstop} = [rectangle, rounded corners, minimum width=3cm, minimum height=1cm,text centered, draw=black, fill=gray!30]
\tikzstyle{process} = [rectangle, minimum width=3cm, minimum height=1cm, text centered, draw=black, fill=orange!30]
\tikzstyle{arrow} = [thick,->,>=stealth]

\sethlcolor{yellow}

\begin{document}

\title{Challenges in Guardrailing Large Language Models for Science}

\author{
    Nishan Pantha$^{1}$, Muthukumaran Ramasubramanian$^{1}$, Iksha Gurung$^{1}$,
    Manil Maskey$^{2}$, Rahul Ramachandran$^{2}$ \\
    $^{1}$Earth Systems Science Center, University of Alabama in Huntsville, Huntsville, AL 35801, USA \\
    $^{2}$Marshall Space Flight Center, NASA, Huntsville, AL 35899, USA
}

\maketitle

\begin{abstract}
The rapid development in large language models (LLMs) has transformed the landscape of natural language processing and understanding (NLP/NLU), offering significant benefits across various domains. However, when applied to scientific research, these powerful models exhibit critical failure modes related to scientific integrity and trustworthiness. Existing general-purpose LLM guardrails are insufficient to address these unique challenges in the scientific domain. We propose a comprehensive taxonomic framework for LLM guardrails encompassing four key dimensions: trustworthiness, ethics \& bias, safety, and legal compliance. Our framework includes structured implementation guidelines for scientific research applications, incorporating white-box, black-box, and gray-box methodologies. This approach specifically addresses critical challenges in scientific LLM deployment, including temporal sensitivity, knowledge contextualization, conflict resolution, and intellectual property protection.

\end{abstract}

\begin{IEEEkeywords}
NLP, NLU, LLM, AI
\end{IEEEkeywords}

\section{Introduction}
The advent of large language models (LLMs) has revolutionized the field of natural language processing and understanding (NLP/NLU) \cite{brown2020language} \cite{radford2018improving} \cite{wei2021finetuned}, leading to numerous applications, particularly in chat systems. Users interact with these systems in both open-ended and close-ended question-answering (QA) modes, leveraging the models' capabilities to generate human-like responses and perform complex language tasks. However, deploying LLMs in real-world applications introduces safety, ethics, and reliability challenges. As a result, extensive research has focused on incorporating LLM guardrails to ensure responsible use and prevent failure modes \cite{bender2021dangers, xu2021bot}.

LLM guardrails are mechanisms designed to enforce safety and various standards in LLM applications by monitoring and controlling user interactions \cite{dong2024safeguarding}. These guardrails operate through both intrinsic model-level constraints and explicit rule-based systems, ensuring LLMs function within predefined principles while validating response structure, type, and quality. This is crucial due to the inherent unpredictability of LLMs, which can generate biased, misleading, or harmful outputs. Effective governance and safety measures are essential to maintain trust in generative AI technologies like these, as they become more integrated into daily applications. Some critical dimensions and properties of LLM guardrails, applicable to a wide range of domains \cite{dong2024safeguarding}, include (but are not limited to):
 \begin{itemize}
    \item Mitigating factual hallucinations
    \item Ensuring fairness in data handling
    \item Enforcing data privacy, confidentiality, and regulatory standards
    \item Enhancing model robustness against adversarial attacks
    \item Detecting and filtering toxic content
    \item Complying with legal and ethical standards
    \item Identifying and Handling out-of-distribution inputs and outputs.
    \item Accurately quantifying and communicating output uncertainty
\end{itemize}

These properties become even more critical in sensitive domains, such as scientific research. High standards of factual accuracy and adherence to content moderation are essential to prevent the generation of inappropriate or misleading responses. The scientific field demands precision, as even minor inaccuracies or biases can have significant consequences \cite{hovy2016social, obermeyer2019dissecting}, from misleading research directions to compromising experimental reproducibility, affecting public trust and the advancement of knowledge \cite{thorp2023chatgpt}. Therefore, developing and implementing systematic techniques to evaluate, analyze, and enhance the performance of LLM guardrails is crucial in these contexts.

Recent years have also seen rapid developments in LLMs and AI systems such as GPT-4 \cite{achiam2023gpt}, Llama-3 \cite{dubey2024llama}, Claude \cite{claude3.5}, Mistral \cite{jiang2024mixtral}, and Gemini \cite{team2023gemini} transforming the landscape of the scientific domain, exhibiting remarkable capabilities in scientific knowledge processing and data discovery, content generation, and data assimilation at scale \cite{bommasani2021opportunities, ai4science2023impact}. These models also have the potential to significantly enhance scientific workflows, from accelerating literature reviews and automating data analysis to aiding in the writing and synthesis of research findings, redefining the bounds of knowledge consumption. \cite{boyko2023interdisciplinary}. However, their positive impact depends on the reliability and accuracy of their outputs. Given the sensitive nature of scientific inquiry, these outputs need to adhere to strict standards.

 Establishing comprehensive guidelines for deploying LLM guardrails in scientific research is therefore crucial. These guidelines should aim to safeguard the integrity of scientific processes, ensuring that the information generated by LLMs is factually accurate, consistent, and aligned with established scientific principles and values. Moreover, the ethical implications of LLM use in science must be thoroughly examined and addressed. Moreover, ethical considerations such as the potential for amplifying biases, privacy concerns, and societal impacts must be carefully addressed \cite{Haltaufderheide_2024, zhang2023ethical}. Effective guidelines should incorporate mechanisms to identify and mitigate these ethical risks, ensuring that the deployment of LLMs in scientific research upholds the highest standards of scientific integrity, fairness, and social responsibility.

This paper explores the key categories and dimensions of LLM guardrails essential for scientific research, including aspects like scientific integrity and biases. Implementing these safeguards requires developing specialized evaluation, analysis, and enhancement techniques. Our aim is to prevent harmful content generation and ensure that LLMs serve as reliable, ethically sound tools in scientific inquiry. By doing so, we seek to utilize LLMs while preserving the accuracy fundamental to the scientific community. In the following sections, we present an overview of aspects of existing LLM guardrail frameworks, emphasizing their roles in supporting trustworthy AI applications in scientific environments. This overview sets the stage for a deeper examination of specific dimensions and challenges in deploying LLM guardrails for scientific applications.
\section{Overview of LLM Guardrails}

Implementing LLM guardrails is crucial to minimize risks, especially when these models are applied in sensitive areas like scientific research. Such guardrails help ensure that LLMs operate within established standards, addressing safety, ethical use, and reliability concerns. In high-stakes fields like science, these measures are vital for maintaining trust and ensuring the responsible use of AI technologies

\subsection{General Dimensions for LLM Guardrails}

Recent surveys, including those by Dong et al. (2024)\cite{dong2024safeguarding}, have highlighted various dimensions for developing effective LLM guardrails. Dong et al. (2024) emphasize the importance of black-box and post-hoc strategies, which involve continuously monitoring and filtering LLM inputs and outputs to safeguard against potential failures. Frameworks such as Llama Guard\cite{inan2023llama}, Nvidia NeMo\cite{rebedea2023nemo}, LMQL \cite{beurer2023prompting}, Guidance \cite{guidance_ai_github}, and Guardrails AI \cite{guardrails_ai_github} are examples of systems designed to enforce these safety measures, ensuring that LLMs operate within established guidelines and standards across various applications.

For example, Llama Guard \cite{inan2023llama} focuses on enhancing human-AI conversation safety by classifying outputs based on user-defined categories. Nvidia NeMo \cite{rebedea2023nemo} employs a more formal approach, using sentence transformers to guide LLMs within strict dialogical boundaries. On the other hand, Guardrails AI provides structure, type, and quality guarantees for LLM outputs, though its applicability is limited to text-based scenarios. The table \ref{table:dong-guardrails-tools} summarizes different properties of these guardrails illustrating their respective strengths and limitations in addressing different aspects of LLM guardrails such as safety, fairness, privacy, robustness, and legality.

\begin{table}[H]
    \centering
    \caption{Abilities among different Guardrails, Dong \textit{et al.} \cite{dong2024safeguarding}}  \begin{tabular}{lcccccc}
        \hline
        & Llama Guard & Nvidia NeMo Guardrails & AI TruLens & Guidance AI & LMQL \\
        \hline
        Hallucination & \checkmark & \checkmark & \checkmark & \checkmark & \checkmark & \checkmark \\
        Fairness & \checkmark & - & \checkmark & \checkmark & - & - \\
        Privacy & - & \checkmark & - & - & - & - \\
        Robustness & - & - & - & - & - & - \\
        Toxicity & \checkmark & \checkmark & \checkmark & \checkmark & \checkmark & \checkmark \\
        Legality & \checkmark & - & - & - & - & - \\
        Out-of-Distribution & - & - & \checkmark & - & - & - \\
        Uncertainty & - & \checkmark & \checkmark & \checkmark & - & - \\
        \hline
    \end{tabular}
    \label{table:dong-guardrails-tools}
\end{table}

Scientific research poses distinct challenges for each of these guardrail aspects. To better understand these challenges, we explore critical dimensions that are particularly relevant to scientific inquiry, such as the dangers of hallucination, the importance of fairness and robustness, and the need for privacy.

\subsubsection{Hallucination}

Hallucinations refer to instances where the LLM generates factually incorrect or nonsensical outputs. This issue is particularly concerning in sensitive applications, such as scientific research, where accuracy is crucial.
For example, \cite{goodman2024ai} notes that in clinical settings even minor hallucinations -- such as adding symptoms like "fever" inaccurately to a patient summary -- can reinforce diagnostic biases and misguide clinical decisions. Similarly, Huang et al. \cite{huang2023survey} examine the broader landscape of hallucinations in LLMs, presenting a detailed taxonomy of contributing factors such as inherent pre-training biases and flawed decoding strategies that can affect reliability in real-world applications.Researchers have explored various approaches to safeguard LLMs against these pitfalls. Techniques like formal verification and adversarial training have been proposed to embed safety checks within the model's architecture, ensuring that outputs adhere to rigorous standards of factual accuracy \cite{dong2024building} \cite{dong2024safeguarding}. Additionally, methods such as ChainPoll \cite{friel2023chainpoll}, SelfCheckGPT \cite{manakul2023selfcheckgpt}, GPTScore \cite{fu2023gptscore}, and G-Eval \cite{liu2023g} have been developed to detect and mitigate hallucinations effectively.

\subsubsection{Fairness}

Fairness in LLMs involves ensuring that the outputs are unbiased and do not perpetuate harmful stereotypes or discrimination. This is particularly challenging given that LLMs are often trained on extensive and diverse datasets encompassing multiple languages, cultures, and ideologies, which can inadvertently embed biases \cite{zhang2023ethical}.
The limitations shown in clinical LLM studies, such as those highlighted in \cite{hager2024evaluation}, demonstrate how these biases can manifest in medical decision-making, potentially impacting vulnerable patient groups. 
Addressing these challenges requires robust bias mitigation strategies during both training and deployment, including techniques like counterfactual data augmentation, debiasing modules \cite{meade2021empirical}, and advanced bias detection and correction mechanisms \cite{zhang2018mitigating}.

\subsubsection{Privacy}

Privacy concerns are particularly relevant in the context of LLMs, as these models are often trained on large datasets that may include sensitive or personally identifiable information (PII). Research has shown that LLMs can inadvertently memorize and reproduce such information, posing risks of data leakage and privacy violations \cite{carlini2021extracting, carlini2019secret, neel2023privacy, staab2023beyond}. Differential privacy and watermarking techniques are commonly employed to mitigate these risks. Tools like ProPILE \cite{kim2024propile} have demonstrated how privacy probes can assess the extent of PII exposure, underscoring the importance of robust privacy-preserving measures in training and deployment to safeguard sensitive data effectively.

\subsubsection{Robustness}

Robustness refers to the model’s ability to maintain performance even when faced with challenging or adversarial inputs. This is critical for preventing the model from being easily manipulated or misled. For instance, LLMs used in chemical compound discovery were found to be susceptible to adversarial attacks that caused the models to suggest invalid molecular structures, compromising the reliability of scientific outputs \cite{wong2024smilesprompting}. Similarly, the incorrect handling of rare scientific terms in biological research led to inconsistent conclusions, highlighting weaknesses in the model's understanding of specialized terminology \cite{zhang2024scientific}. Adversarial training \cite{zhang2018mitigating, jain2023baseline} and robustness testing \cite{perez2022red, wang2023robustness}  are common approaches to enhance the resilience of LLMs.

\subsubsection{Toxicity and Legality}

LLMs must be safeguarded against generating toxic or illegal content, such as hate speech or misinformation \cite{bender2021dangers}. For example, in medical research, there have been cases where LLMs generated misleading treatment suggestions, potentially causing harm to patients if used unchecked \cite{chang2024red}. In legal contexts, LLMs have also produced outputs that inadvertently violated data protection regulations, leading to privacy breaches and regulatory issues \cite{staab2023beyond}. \cite{zhang2023right} examines such challenges of ensuring LLMs comply with data protection laws, highlighting instances where models have memorized and leaked sensitive information. Content moderation techniques, including the use of toxicity classifiers and red-teaming exercises, help to ensure that LLM outputs remain within acceptable ethical and legal boundaries \cite{gehman2020realtoxicityprompts, zhang2022opt, liao2023ai}.

\subsubsection{Uncertainty and Explainability}

Uncertainty quantification \cite{zhu2023calibration, sankararaman2022bayesformer} is crucial for assessing the confidence of LLM outputs, particularly in scientific and decision-making contexts. For example, in a real-world application involving LLMs for drug discovery, the lack of explainability led to misunderstandings about the efficacy of proposed compounds, resulting in wasted research efforts and resources \cite{wysocka2023large, zheng2024large}. Similarly, in climate modeling, unexplained discrepancies in LLM-generated forecasts led to incorrect conclusions, affecting policy decisions \cite{bulian2023assessing, fore-etal-2024-unlearning, kraus2023enhancing}. Techniques like self-consistency \cite{ahmed2023better} and chain-of-thought (CoT) prompting \cite{wei2022chain} are used to improve the transparency and reliability of LLM reasoning processes \cite{jiang2021can}. Tools such as LIME \cite{lime} and SHAP \cite{NIPS2017_7062} have also been adopted to help explain model outputs, providing a deeper understanding of the decision-making pathways of LLMs in scientific contexts.

Building on these foundations, Dong et al. (2024) \cite{dong2024building} discuss the challenges in constructing comprehensive guardrails, particularly when balancing multiple, sometimes conflicting requirements—such as fairness, privacy, and robustness. These challenges are compounded by the fact that the requirements for these guardrails often interact in complex ways, as highlighted by Raji and Buolamwini (2019) \cite{raji2019actionable} in their discussion on the impact of biased datasets on AI outcomes, and Carlini et al. \cite{carlini2019secret, carlini2021extracting}, who demonstrated how LLMs could inadvertently memorize and expose private information. \cite{mittal2023responsible, chen2023privacy} also emphasize the trade-offs between maintaining privacy and achieving fairness in machine learning models. They highlight the difficulty of establishing these guardrails due to the diverse nature of these requirements across different applications. A multidisciplinary approach, integrating both symbolic and learning-based methods, is proposed to address these challenges, ensuring that guardrails can adapt to the evolving capabilities of LLMs while maintaining rigorous safety standards. Furthermore, \cite{zarifzadeh2024lowcost, manzonelli2024membership, wen2024membership} provide insights into privacy vulnerabilities through membership inference attacks, which underscore the need for robust privacy guardrails in scientific contexts.

While existing LLM guardrail frameworks provide essential safety measures, the scientific domain presents distinct requirements that extend beyond conventional safeguards. Scientific applications demand additional layers of verification, reproducibility, and methodological rigor that current frameworks do not fully address. The following section examines these domain-specific challenges and introduces novel guardrail dimensions crucial for maintaining scientific integrity in LLM-assisted research.

\subsection{Challenges of Implementing LLM Guardrails for Science}

While LLMs present significant advantages across various domains, their deployment within scientific contexts introduces a unique set of challenges that must be addressed to maintain scientific integrity. These challenges primarily revolve around issues of scientific integrity, reliability, ethical considerations, and legal compliance. Scientific research demands precise and trustworthy information, making the mitigation of errors, biases, and unethical outputs crucial.

 The importance of enhancing LLMs to prevent harmful outputs is further underscored by Tang et al. (2024)\cite{tang2024prioritizing}. They examine the specific vulnerabilities of LLM-based agents in scientific domains, emphasizing the need for a triadic framework involving human regulation, agent alignment, and environmental feedback. This framework aims to mitigate risks such as factual errors, jailbreak attacks, and the misuse of scientific information. The authors argue that safeguarding efforts should prioritize risk control over the autonomous capabilities of LLMs, particularly in high-stakes scientific environments. In the context of scientific research, biases can differ significantly from those typically encountered in general-purpose LLMs. While general-domain biases often involve demographic-based issues, such as gender, race, or cultural stereotypes, biases in the scientific domain tend to focus on inaccuracies related to research quality, methodological errors, over-representation of certain hypotheses, and institutional or publication and positivity biases. These scientific biases can result in the disproportionate representation of particular theories, flawed conclusions due to biased training data, or even reinforcing historically dominant research paradigms.

Addressing these challenges requires implementing domain-specific solutions on top of general guardrails to ensure that LLMs support the advancement of reliable and ethical scientific research. Some of these specific challenges include:

\subsubsection{Hallucination Mitigation}
One of the most significant challenges is preventing hallucinations — instances where LLMs generate factually incorrect or misleading outputs. This is an issue in scientific contexts where the accuracy of information is critical. These hallucinations can either be open-domain, where models make false general claims easily verifiable against reliable sources, or closed-domain where the models deviate from the provided context or reference text\cite{friel2023chainpoll}.
Such inaccuracies in scientific contexts can have severe consequences, undermining the integrity and credibility of research outputs. Current approaches to hallucination prevention, such as formal verification and adversarial training, may not fully integrate with domain-specific knowledge bases \cite{gururangan2020don, li2024enhancing, cao2023learn, tonmoy2024comprehensive}, which are crucial for ensuring scientific accuracy.

\subsubsection{Temporal Relevancy}
Another critical challenge is ensuring that LLM-generated content remains up-to-date with the latest scientific findings. Scientific research is a dynamic field, and outdated information can lead to erroneous conclusions, negatively impacting the credibility of the research. Existing frameworks may lack mechanisms to ensure the time relevancy of LLM outputs, limiting their applicability in fast-evolving scientific domains.

\subsubsection{Conflict Identification and Resolutiuon}
Handling conflicting research results is a complex issue that general LLM approaches may not adequately address. In fields where research findings frequently evolve or contradict previous studies, LLMs must be capable of reconciling these conflicts to provide reliable and coherent outputs. However, current guardrails may not possess the necessary sophistication to manage such complexities effectively

\subsubsection{Consistency}
Maintaining consistency across outputs is crucial in scientific research. Inconsistent outputs can undermine the validity of research findings, leading to a lack of coherence and reliability in the generated content. Ensuring consistency also helps in building a reliable knowledge base that other researchers can trust and build upon. Moreover, maintaining consistency reduces the likelihood of conflicting interpretations, thereby safeguarding the integrity of the overall scientific inquiry.

\subsubsection{Attribution}
Ensuring proper attribution and citation is equally important in scientific research. Inadequate citation practices can lead to plagiarism and intellectual dishonesty, which can severely compromise the integrity of the research. Inconsistent outputs can undermine the validity of research, while inadequate citation practices can lead to plagiarism and intellectual dishonesty. General LLM guardrails may not enforce the rigorous standards required by scientific journals, posing risks to the integrity of the research.

\subsubsection{Explainability and Transparency}
 In scientific research, the explainability of LLM outputs is essential for ensuring transparency and traceability. Researchers need to understand the reasoning process behind LLM-generated content to verify its alignment with established scientific principles. However, general-purpose guardrails often provide only superficial reasoning chains, lacking the depth required for rigorous scientific validation.

 \subsubsection{Ethical and Legal Considerations}
 The ethical and legal challenges associated with LLMs in scientific research are significant. These include ensuring fairness, mitigating biases, protecting privacy, and adhering to legal standards such as intellectual property rights. General LLM guardrails may not fully address these challenges, particularly in contexts like clinical research, where the stakes and requirements are high.
 
 To address these challenges, He et al. \cite{he2023control} proposepossible enhancements to these guardrails that could be effective in scientific domains. They propose SciGuard, a system specifically designed to control the misuse risks of AI models in science. SciGuard serves as a mediator between users and AI models, implementing ethical and safety standards customizable to different domains. The growing prevalence of AI applications in science brings escalating concerns about their potential misapplication. It highlights the urgency for systems like SciGuard that can prevent unintended harm even when AI is used with good intentions. They further highlight that the dynamic nature of risks in scientific AI necessitates continuous monitoring and reassessment to ensure the effectiveness and relevance of risk mitigation strategies, making it imperative that future development focuses on building safeguarded scientific AI systems that align with both ethical standards and scientific integrity.

 While prior studies have explored general-purpose guardrails for LLMs \cite{dong2024safeguarding}, there remains a significant gap in addressing the unique challenges specific to scientific research. To bridge this gap, we reformulate the guardrails within a different framework that extends traditional guardrail dimensions, incorporating aspects relevant to science such as time sensitivity, knowledge contextualization, and conflict resolution. Our contributions are two-fold:
\begin{itemize}
    \item \textbf{Expansion of LLM guardrail dimensions to support applications in the scientific domain}: We introduce a categorization framework that expands the traditional guardrail dimensions to include those uniquely pertinent to scientific research.

    \item \textbf{Implementation strategies}: We provide actionable strategies for implementing these guardrails using a hybrid of white-box, black-box, and gray-box approaches, motivated by a similar approach taken by Dong et al \cite{dong2024safeguarding} for preventing LLM attacks.

\end{itemize}

In the next section, we propose specific dimensions for implementing effective LLM guardrails tailored to scientific research.
\section{Dimensions for Guardrails for Science}

In the previous section, we identified various gaps in applying general LLM guardrails to the scientific domain, such as the lack of domain-specific adaptation, insufficient temporal relevance, and challenges in handling multidisciplinary data. Here we propose a set of tailored areas that address and encompass those properties within specific scientific needs. These dimensions are categorized under four main areas: \textit{Trustworthiness}, \textit{Ethics \& Bias}, \textit{Safety}, and \textit{Legal}. Each category is further subdivided into critical areas for improvement, offering a roadmap for applying LLM guardrails more effectively within scientific contexts.

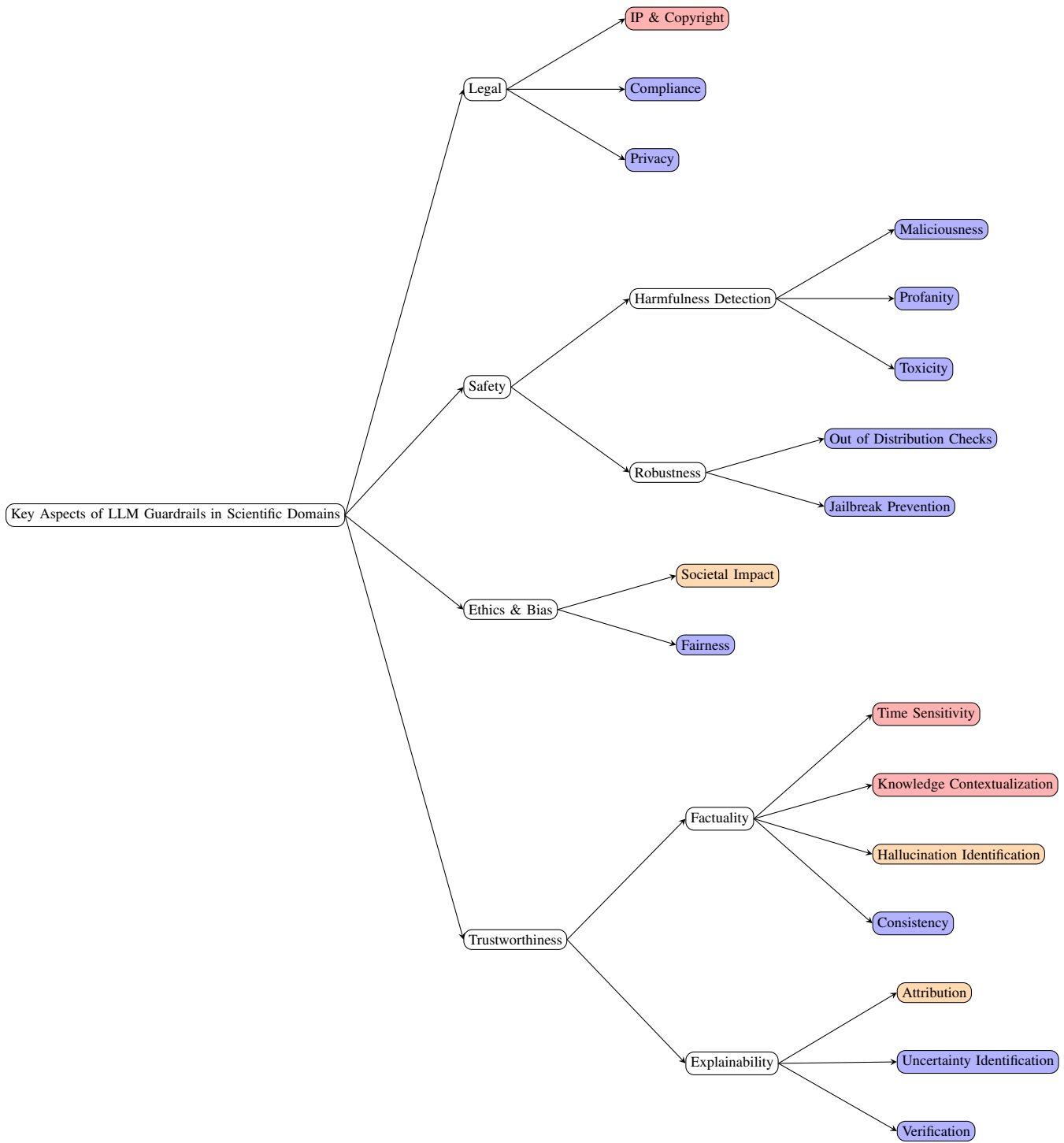
\begin{figure}[!h]
\centering
\begin{forest}
for tree={
    grow=east,
    draw,
    rounded corners,
    edge={-stealth},
    l sep=2cm, 
    s sep=0.8cm, 
    parent anchor=east,
    child anchor=west,
    calign=center,
    calign primary angle=0,
    calign secondary angle=0,
    fit=band,
    scale=.7,
    anchor=base west,
}
[Key Aspects of LLM Guardrails in Scientific Domains,
    [Trustworthiness,
        [Explainability,
            [Verification, fill=blue!30]
            [Uncertainty Identification, fill=blue!30]
            [Attribution, fill=orange!30]
        ]
        [Factuality, name=factuality,
            [Consistency, fill=blue!30]
            [Hallucination Identification, fill=orange!30]
            [Knowledge Contextualization, fill=red!30]
            [Time Sensitivity, fill=red!30]
        ]
    ]
    [Ethics \& Bias,
        [Fairness, fill=blue!30]
        [Societal Impact, fill=orange!30]
    ]
    [Safety,
        [Robustness,
            [Jailbreak Prevention, fill=blue!30]
            [Out of Distribution Checks, fill=blue!30]
        ]
        [Harmfulness Detection,
            [Toxicity, fill=blue!30]
            [Profanity, fill=blue!30]
            [Maliciousness, fill=blue!30]
        ]
    ]
    [Legal, name=legal,
        [Privacy, fill=blue!30]
        [Compliance, fill=blue!30]
        [IP \& Copyright, fill=red!30]
    ]
]
\end{forest}
\caption{Key Aspects of LLM Guardrails in Scientific Domains}
\label{forest:llm-guardrails-science-tree}
\end{figure}

The dimensions are represented in a structured framework, illustrated in Figure \ref{forest:llm-guardrails-science-tree}, and categorized by the level of adaptation required. We use a color-coding scheme -- blue, orange, red, and uncolored -- to indicate the extent of modification needed for existing LLM guardrails to be applied to scientific research. Blue boxes signify areas where minimal adaptation is necessary, orange boxes denote dimensions that require some refinement, red boxes represent areas where significant development is required, and uncolored boxes represent a categorization that indicates areas already addressed by LLMs across multiple domains, rather than specific, well-defined guardrails.

We expand each category, highlighting key dimensions and explaining their importance in scientific research. Each of these dimensions has been chosen for detailed exploration based on its critical relevance to the unique challenges presented in scientific contexts. While LLMs can contribute broadly across various domains, these particular dimensions require specialized adaptations to meet the high standards of scientific research. For example, \textit{Compliance} ensures that research adheres to institutional, legal, and ethical guidelines \cite{dubois2016compliance, watkins2023guidance, resnik2024ethics}, while \textit{Attribution} guarantees proper credit is given to original sources, a critical aspect of maintaining scientific integrity \cite{macrina2014scientific, national2017fostering, huang2023citation}. \textit{Time Sensitivity} is important in fields where the timeliness of information can affect research outcomes \cite{mousavi2024dyknowdynamically}, and \textit{Knowledge Contextualization} ensures that the vast amount of information processed by LLMs is relevant and accurate within the specific scientific context it is applied \cite{ning2024user}. We deliberately chose not to expand on every dimension in detail, as some are well-understood and already have robust guardrails in place that apply broadly across domains. By narrowing our focus to these specific dimensions, we aim to provide deeper insights into areas where LLMs must be particularly adapted or enhanced to meaningfully contribute to advancing scientific knowledge while maintaining the highest standards of reliability and integrity.

\subsection{Blue Boxes}

The blue boxes represent the dimensions of LLM guardrails that can be adapted for scientific use with minimal modification. These are established best practices that can transition smoothly into the scientific context, requiring only slight adjustments. The goal here is to leverage these existing guardrails to enhance LLMs for the specific needs of scientific research, ensuring their outputs are dependable and contextually appropriate. The following dimensions outline the key areas where LLM guardrails must be implemented to ensure they are suitable for scientific research. These dimensions aim to enhance the reliability, safety, and ethical compliance of LLM outputs

\begin{itemize}
    \item \textbf{Verification:} 
    Ensuring that LLM outputs can be verified against known, established sources.
    
    \item \textbf{Uncertainty Identification:} 
    Mechanisms for identifying uncertainty in LLM outputs.
    
    \item \textbf{Consistency:} 
    Maintaining consistency in information, particularly concerning scientific data.
    
    \item \textbf{Fairness:} 
    Addressing fairness concerns to prevent biased or prejudiced outputs.
    
    \item \textbf{Robustness:} 
    Strengthening LLMs against potential failures, including jailbreak prevention and out-of-distribution checks.
    
    \item \textbf{Harmfulness Detection:} 
    Detecting harmful outputs related to toxicity, profanity, and maliciousness.
    
    \item \textbf{Privacy:} 
    Ensuring LLMs comply with privacy laws and safeguard personal information.

    \item \textbf{Compliance:} 
    Ensuring that LLMs adhere to legal, ethical, and institutional guidelines. 
\end{itemize}

\subsubsection{Compliance}

The \textbf{Compliance} dimension ensures that the identification of similar research work and relevant grants adhere to legal, ethical, and institutional guidelines, while also promoting trustworthiness in scientific research. This functionality is critical for optimizing resource allocation, avoiding research duplication, and fostering collaboration. This also focuses on developing sophisticated tools that can automatically search and compare existing literature, grant opportunities, and ongoing research projects, thereby aiding researchers in navigating the vast and ever-expanding body of scientific knowledge. For instance, an LLM could analyze a researcher’s proposal and automatically suggest related studies or previous work in the same area, helping to ensure that the new research builds on existing knowledge rather than inadvertently replicating it. No guardrail dimensions exist yet that address this problem. While there has been some work \cite{zhu2023novel} on matching grants in a specific domain, there has yet to be proper discipline on the implementation in a wider scientific domain.

Similarly, it also ensures that the works follow publication guidelines and scientific values. Identifying related grants is particularly valuable for researchers seeking funding opportunities that align with their projects. By scanning grant databases and matching them with the researcher's area of study, LLMs can suggest relevant funding opportunities, streamlining the grant application process and increasing the chances of securing support. This functionality saves time and opens up potential collaborations by identifying research groups working on similar problems.

The development of these capabilities would require LLMs to perform comprehensive searches and analyses, cross-referencing multiple sources of academic literature and funding databases. Advanced natural language processing (NLP) techniques would be necessary to accurately parse and understand the nuanced details of research proposals and grant descriptions, ensuring that the recommendations \cite{wu24survey} made are highly relevant and useful to the researchers.

Incorporating these search and comparison tools into LLMs would empower researchers to make more informed decisions, enhance the efficiency of the research process, and foster a more collaborative scientific community. These tools represent a significant step toward integrating AI more deeply into the research lifecycle, supporting the broader goals of innovation and knowledge advancement in the scientific community. \cite{floridi2022unified}

\subsection{Orange boxes}

The orange boxes represent existing dimensions but require significant refinement to meet the unique demands of scientific integrity, accountability, and precision. These dimensions are crucial in ensuring that LLMs provide accurate outputs, uphold ethical standards, and promote contextual understanding within scientific disciplines.

\begin{itemize}
    \item \textbf{Attribution:} 
    Ensuring proper credit to original sources to maintain academic integrity.  
    
    \item \textbf{Hallucination Identification:} 
    The ability to recognize when LLMs provide inaccurate or fabricated information.
    
    \item \textbf{Societal Impact:} 
    Considering the broader societal effects of LLM-generated outputs.
    
\end{itemize}

\subsubsection{Attribution}
Proper attribution is essential in scientific research to acknowledge the original sources, avoid plagiarism, and maintain academic integrity. LLMs may inadvertently generate content that closely mirrors existing literature without appropriate citations, raising concerns about intellectual dishonesty and transparency of LLM-generated content. To address this, LLMs must be incorporated with mechanisms for accurate attribution \cite{huang2023citation}, such as generating citations in the correct academic formats, implementing source-tracking capabilities, and integrating plagiarism detection algorithms. \cite{gao2022rarr} \cite{bohnet2022attributed}

Challenges include the complexity of source materials, knowledge cutoffs \cite{zhu2024llm} leading to outdated citations, and variability in citation styles across disciplines. Strategies to improve attribution involve integrating LLMs with citation databases like PubMed or CrossRef, employing retrieval-augmented generation techniques \cite{lewis2020retrieval} to access the latest publications, and standardizing citation practices within the models. By prioritizing accurate attribution, LLMs can enhance the trustworthiness of their outputs, uphold ethical standards, and facilitate verification of information, thereby fostering a more transparent and reliable scientific community.

\subsection{Red boxes} 

The red boxes represent dimensions where current LLM guardrails are either underdeveloped to meet scientific standards or non-existent in the general literature. These dimensions remain underdeveloped precisely because they address challenges unique to scientific discourse—such as truthfulness, accuracy, and reproducibility—which rarely arise in general-purpose applications. For scientific research, certain attributes require a significant focus due to the unique challenges they present. For example, \textit{Time Sensitivity} is essential because scientific knowledge evolves rapidly, and LLMs must provide information that reflects the latest developments to remain relevant. \textit{Knowledge Contextualization} is equally crucial, as scientific recommendations must be tailored to specific fields, regions, or study conditions to be genuinely useful. By expanding on these dimensions, we aim to ensure that LLMs are better equipped to handle the complexity and dynamism of scientific research. These areas are crucial for the future development of LLMs tailored to scientific research.

\subsubsection{Time Sensitivity}

LLMs are trained on static datasets with a specific knowledge cutoff date and lack mechanisms for real-time knowledge updates \cite{zhu2024llm} \cite{mousavi2024dyknowdynamically}. This limitation presents several critical issues with relying on LLMs for scientific applications:

\begin{itemize}
    \item \textbf{Static Knowledge and Rapid Obsolescence}: LLMs are "frozen" at their last point of training, unable to incorporate or reflect the most recent advancements unless explicitly retrained. \cite{zhu2024llm, mousavi2024dyknowdynamically} In rapidly evolving scientific fields like epidemiology, biotechnology, or climate science, research findings and policy changes occur frequently, rendering information obsolete within months or weeks. This static nature is especially problematic in domains where timely, accurate information is essential for decision-making, such as public health, finance, or crisis management \cite{roberto2024updating}. LLMs cannot respond to breaking news or incorporate real-time data about stock market fluctuations, emerging viral outbreaks, or sudden regulatory changes, limiting their utility in environments that demand agility and current knowledge.

    \item \textbf{Consequences of Outdated Information and False Confidence}: LLMs often present outdated information with high confidence, lacking built-in mechanisms to indicate that their knowledge may be outdated. This makes it difficult for users to discern whether the information provided is still accurate or relevant. For example, a scientist seeking advice on the latest vaccine technologies may receive authoritative-sounding responses that reflect research no longer considered cutting-edge. Relying on outdated information can lead to significant consequences, including the spread of misinformation, misallocation of resources, and ineffective strategies \cite{vykopal2023disinformation}. In fields like public health, this could result in promoting vaccines less effective against new viral strains, leading to increased infection rates and unnecessary strain on healthcare systems. The model's high confidence in outdated data exacerbates the risk of misguided conclusions or recommendations, as users may not be alerted to the possibility that newer research has changed the consensus.

\end{itemize}

Therefore, as scientific knowledge is dynamic and constantly evolving, LLMs must provide information that is not only accurate but also temporally relevant. The challenge is that many LLMs are trained on datasets that may become outdated \cite{zhu2024llm} \cite{mousavi2024dyknowdynamically}, leading to the generation of content that no longer reflects the current state of scientific understanding. To address this, models need to be equipped with mechanisms to verify the temporal validity of information by cross-referencing against real-time data sources. This could involve integrating LLMs with updated knowledge bases or using retrieval-augmented generation (RAG) \cite{lewis2020retrieval} techniques to pull in current information as part of the response generation process. 

\subsubsection{Knowledge Contextualization}

Similar to time sensitivity, LLM responses must adapt their responses to the specific context in which information is applied. In scientific research, the same data can produce varying conclusions depending on the region, field of study, or unique circumstances of the user’s inquiry. \textbf{Knowledge contextualization} ensures that LLMs take these variations into account, allowing for more nuanced, tailored responses. This means recognizing disciplinary differences, adjusting recommendations based on the user’s level of expertise \cite{ning2024user}, and considering geographical or environmental factors that may influence how a particular piece of knowledge should be applied. By understanding the context, LLMs can offer outputs that are not only factually correct but also pragmatically relevant to the specific research or situation at hand.

For example, in agricultural research, a general recommendation on farming best practices may be valid for a temperate region but completely unsuitable for arid climates. A lack of contextualization could result in crop damage or environmental harm. LLMs need the ability to adapt information dynamically to reflect these unique local conditions, which often requires an understanding of diverse disciplines, localized expertise, and complex interdependencies within the context of a specific inquiry. Below are key challenges that emerge when LLMs fail to properly contextualize knowledge:

\begin{itemize}

    \item \textbf{Overgeneralization:}
    LLMs often provide generalized responses that fail to account for local variations or specific situational needs. This can lead to recommendations that are ineffective or even harmful when applied in contexts with unique environmental, demographic, or infrastructural factors. Overgeneralization risks poor decision-making and unintended negative consequences, particularly in domains like agriculture and healthcare, where local conditions are critical for successful outcomes.
    
    \item \textbf{Failure to Integrate Multidisciplinary Knowledge:}
    LLMs often struggle to synthesize information from multiple disciplines, providing narrow, discipline-specific responses. This can lead to incomplete or unbalanced recommendations, particularly in complex fields like sustainable farming, where solutions require integrating knowledge from climatology, ecology, economics, and soil science. The lack of multidisciplinary integration may result in decisions that optimize one aspect while neglecting others, leading to suboptimal or harmful outcomes, even if the query to LLM might seem benign. One potential mitigation strategy could be to adapt LLMs to use tools that incorporate these disciplines. \cite{lu2024chameleon}
    
    \item \textbf{Lack of Adaptive Reasoning:}
    LLMs often lack the ability to adjust their responses to the specific context of a problem, such as local economic, environmental, or social factors. Adaptive reasoning \cite{chentoward} \cite{DBLP:conf/emnlp/AggarwalY23} requires weighing multiple, sometimes conflicting variables, which LLMs struggle to do effectively. Without this ability, they may offer one-size-fits-all recommendations that overlook critical nuances, leading to poor decision-making and undesirable outcomes in areas like public policy, healthcare, or disaster management.
    
    \item \textbf{Positivity Bias}: LLMs tend to exhibit a positivity bias \cite{tjuatja2024llms, bai2024measuring}, often providing overly optimistic responses that may not accurately reflect the underlying data or context. This bias can lead to an unrealistic portrayal of outcomes or underestimation of risks, particularly in fields like healthcare or environmental management. Proper contextualization of facts and more nuanced evaluations are necessary to ensure that responses are balanced and reflect the true range of potential outcomes.

\end{itemize}

Additionally, it is common to encounter studies with differing conclusions involving the synthesis of conflicting data and perspectives. Current LLMs may struggle to handle these conflicts effectively \cite{xu2024knowledge}, leading to oversimplified or biased outputs. LLMs need to be able to identify and appropriately manage these conflicts, rather than presenting conflicting results as equally valid without distinction. Therefore, it is necessary to develop frameworks that allow LLMs to recognize, reconcile, and accurately represent conflicting findings in scientific literature. This would involve integrating advanced reasoning algorithms like chain-of-thoughts \cite{wei2022chain}, chain-of-verification \cite{dhuliawala2023chain}, verify-and-edit \cite{zhao2023verifyandedit} that can weigh evidence, evaluate the credibility of sources, and present balanced conclusions. Furthermore, solutions could involve developing models that can prioritize information based on the consensus within the scientific community or highlight the divergence in findings to guide users in making informed decisions. Conflict resolution mechanisms within LLMs could be employed to manage these situations effectively \cite{liu-etal-2024-untangle} \cite{chern2023factool}.

\subsubsection{IP \& Copyright}

Intellectual property (IP) and copyright play a fundamental role in scientific research, where the originality and ownership of ideas, data, and innovations are crucial \cite{sobel2017artificial}. This dimension focuses on developing mechanisms within LLMs that can accurately identify, track, and manage intellectual property and copyrighted content, ensuring that the outputs generated respect the proprietary rights of researchers, institutions, and other stakeholders. By integrating these capabilities, LLMs can uphold legal and ethical standards, maintaining the integrity of the scientific process while protecting the rights of individuals and organizations.

As LLMs generate content, it is crucial to recognize and label portions of text that may be novel, patentable, or otherwise protected under intellectual property laws \cite{sobel2017artificial}. This capability is essential not only for protecting IP rights but also for guiding researchers and organizations in managing and utilizing these outputs. For instance, if an LLM generates an innovative idea or a unique piece of content, the system should be able to flag this as potentially patentable, alerting users to the need for further steps to secure IP protection \cite{chesterman2024good}.

Moreover, ensuring that attribution of sources is a vital component of IP and copyright management. LLMs should be equipped with citation and attribution mechanisms to accurately acknowledge the origins of data, ideas, or prior work referenced in generated content. This preserves academic integrity and mitigates legal risks related to IP infringement or unauthorized use of proprietary information. By doing so, the system can identify instances where the output may require licensing or other legal compliance before it can be used commercially \cite{abdikhakimov2023unraveling}.

This proposed categorization of LLM guardrails addresses the complex challenges of using LLMs in scientific applications that emphasize trust, ethical responsibility, safety, and legal compliance. We focused primarily on dimensions highlighted by the red boxes, which represent areas where current LLM guardrails are insufficient for the scientific domain. In the following section, we outline implementation strategies for enhancing these guardrails, categorized as white-box, black-box, and gray-box approaches to meet the unique demands of the scientific domain.

\section{Implementation Strategies}

To ensure that the proposed framework for LLM guardrails in the scientific domain is effectively implemented, we adapt the categorization of guardrail-attack types from Dong et al. \cite{dong2024safeguarding} into our approach for guardrail implementation: \textbf{White-box}, \textbf{Black-box}, and \textbf{Gray-box}. While Dong et al. originally used these categories to classify attack vectors based on model access levels, we extend this paradigm to systematically organize guardrail implementation strategies. In their work, white-box attacks assume full visibility into the model’s parameters, black-box attacks restrict access to observing model outputs, and gray-box attacks offer partial access, such as to training data. We adapted this framework to describe corresponding strategies for enforcing guardrails based on the level of access and intervention available within the model. These strategies form a comprehensive framework to address challenges related to trustworthiness, safety, ethics, and legal compliance within the specific context of scientific research, in various levels of effort.

\subsection{White-Box Approaches}

White-box approaches leverage direct access to and modification of LLM internal architecture and its parameters. This allows for precise control over the model’s behavior, providing the ability to enforce guardrails from within the system itself. Techniques such as fine-tuning, model optimization, and formal verification allow developers to adjust and improve the model’s outputs in a controlled and trustworthy manner.

\textbf{Model Fine-Tuning:} This approach involves adjusting the weights and parameters of the LLM through additional training data, allowing developers to refine the model’s behavior. In the context of scientific research, it helps reduce biases and enhance the accuracy of the model’s output, especially for domain-specific tasks like handling complex datasets or ensuring factual correctness. \cite{ouyang2022training} \cite{rafailov2024direct}

\textbf{Architectural Modifications:} This approach involves modifying the underlying architecture of the LLM \cite{mitchell2021fast}, such as introducing new layers or mechanisms (e.g. memory systems \cite{wu2022memorizing}, mixture-of-experts \cite{du2022glam}), enables more robust internal guardrails. These modifications can optimize the model for key scientific challenges such as consistency, uncertainty quantification, and knowledge contextualization. \cite{pmlr-v162-borgeaud22a}

\textbf{Bias Mitigation Techniques:} These techniques involve embedding fairness and bias mitigation \cite{meade2021empirical} \cite{zhang2018mitigating} directly within the model, ensuring that the LLM produces equitable outcomes across different datasets. This is crucial for scientific domains where unbiased and reliable results are essential.

\textbf{Formal Verification:} This technique involves proving that the model’s behavior conforms to predefined standards or rules, ensuring trustworthiness by verifying specific properties of the system. For example, formal verification can ensure that an LLM does not produce harmful or misleading results. \cite{huang2024survey}

\textbf{Adversarial Training and Robustness:} In scientific domains where accuracy and robustness are critical, adversarial training helps the model learn to resist potential attacks or manipulations, ensuring that outputs remain reliable even under challenging input conditions. \cite{zhang2018mitigating} \cite{perez2022red}

White-box strategies are particularly effective for scenarios requiring high levels of control and precision, making them ideal for working with sensitive scientific data that demands rigorous internal safety and trustworthiness mechanisms. While these methods allow for superior control by enabling direct modification of the LLM's internal workings, they are often harder to implement due to their complexity, requiring substantial resources and expertise. However, they introduce significant complexity, largely due to the resource-intensive nature of training and fine-tuning, the need for domain-specific expertise, and the challenge of acquiring large, high-quality datasets. For instance, reinforcement learning from human feedback (RLHF) \cite{ouyang2022training} fine-tunes \cite{bai2022training} the LLM based on human feedback to align the model with specific scientific goals, improving its trustworthiness. Yet, RLHF requires substantial resources, curated datasets, and continuous human oversight, adding to the complexity of implementation. Moreover, scientific data's diversity and complexity, along with the need to regularly update models with new findings, complicate the maintenance of these internal guardrails. Thus, while white-box methods provide better control and precision, their resource demands and complexity may limit practical downstream implementation, especially for smaller organizations with constrained resources.

\subsection{Black-Box Approaches}

In contrast to white-box methods, black-box approaches rely on external mechanisms to enforce guardrails without modifying the LLM’s internal structure. These approaches monitor, filter, and adjust the outputs of the model from the outside, making them effective when direct access to the model is not possible or when the model's internal workings must remain unchanged.

\textbf{Output Filtering:} This approach uses post-processing tools \cite{inan2023llama} \cite{guardrails_ai_github} to monitor and filter the outputs of the LLM. It ensures that generated content aligns with ethical guidelines, scientific standards, and legal requirements. Output filtering is particularly effective for managing issues such as toxicity and hallucination identification in a scientific context, ensuring the content remains reliable. \cite{gehman2020realtoxicityprompts} \cite{xu2020recipes}

\textbf{Rule-Based Post-Processing:} This method applies predefined rules to the output and identifies and controls any content that violates those rules \cite{rebedea2023nemo}. For example, rules could be set to ensure that the model’s predictions adhere to specific scientific norms or avoid controversial statements. This helps in ensuring that the model's outputs align with ethical standards such as fairness and societal impact.

\textbf{External Fact-Checking:} Scientific contexts often require high factual accuracy. By integrating external fact-checking systems \cite{rebedea2023nemo} \cite{google_factcheck_2024}, the black-box approach allows for real-time validation of model outputs \cite{manakul2023selfcheckgpt}, ensuring that incorrect or misleading information is flagged or corrected before being disseminated. \cite{vykopal2024generativelargelanguagemodels} \cite{wei2021finetuned}

\textbf{Content Moderation:} In domains where sensitive information is involved, content moderation systems \cite{lai2022human} \cite{rebedea2023nemo}  can help prevent the dissemination of harmful, inappropriate, or legally problematic outputs, such as biased or inaccurate scientific data. This ensures that the LLM maintains a high standard of safety and reliability. \cite{kumar2024watch}

\textbf{Adversarial Input Detection:} Black-box approaches can also involve detecting and mitigating adversarial inputs that could exploit vulnerabilities in the LLM. This approach ensures that the model maintains robustness even in challenging scenarios by flagging these inputs before they affect the output. \cite{zheng2023judging}

Black-box strategies are ideal for integrating external guardrails with pre-existing LLMs, especially in scientific research where ethical, legal, and factual considerations are critical. A common example of this approach is output filtering, where the LLM processes user input, and the generated output is passed through external guardrails, such as content filters to remove harmful language, PII redaction for privacy compliance, word filters to block specific terms, and denied topics to prevent restricted content. Additionally, a contextual grounding check ensures factual accuracy and relevance. Encoder-based models \cite{devlin2018bert, liu2019roberta}, often used for classification tasks, effectively classify the generated outputs to ensure they adhere to ethical and factual standards \cite{inan2023llama}. They can be used to label outputs for compliance, detect sensitive topics, or evaluate whether the generation aligns with pre-defined guidelines, thereby supporting responsible LLM outputs. These strategies enforce responsible AI policies without modifying the model's internal structure, making them adaptable, cost-effective, and easy to implement \cite{cheng-etal-2024-black}. However, their effectiveness depends heavily on the quality of external rules and filters, which can lead to challenges in managing complex or nuanced scenarios. Furthermore, black-box methods can introduce latency in processing and limit control over deeper model behavior, particularly in cases where more precise intervention is required \cite{dong2024building}.

\subsection{Gray-Box Approaches}

Gray-box approaches offer a middle ground that combines the control and precision of white-box methods with the adaptability of black-box methods, providing a hybrid solution that allows for both internal modifications and external interventions. These strategies provide flexibility, making them particularly useful for managing dynamic, evolving challenges in scientific research where internal model adjustments and external verifications are both necessary.

\begin{table}[H]
\centering
\footnotesize
\begin{adjustbox}{max width=\textwidth}
\begin{tabular}{|c|p{5.7cm}|*{12}{c|}}
\hline

\multirow{3}{*}{\textbf{\large{Type}}} & \multirow{3}{*}{\textbf{\large{Strategy}}} & \multicolumn{7}{c|}{\textbf{\large{Trustworthiness}}} & \multicolumn{2}{c|}{\textbf{\large{Ethics \& Bias}}} & \multicolumn{2}{c|}{\textbf{\large{Safety}}} & \multicolumn{1}{c|}{\textbf{\large{Legal}}} \\ \cline{3-14}
 &  & \rotatebox{90}{\large{\textbf{Verification}}} & \rotatebox{90}{\large{\textbf{Uncertainty Identification}}} & \rotatebox{90}{\large{\textbf{Attribution}}} & \rotatebox{90}{\large{\textbf{Consistency}}} & \rotatebox{90}{\large{\textbf{Hallucination Identification}}} & \rotatebox{90}{\large{\textbf{Knowledge Contextualization}}} & \rotatebox{90}{\large{\textbf{Time Sensitivity}}} & \rotatebox{90}{\large{\textbf{Fairness}}} & \rotatebox{90}{\large{\textbf{Societal Impact}}} & \rotatebox{90}{\large{\textbf{Robustness}}} & \rotatebox{90}{\large{\textbf{Harmfulness Detection}}} & \rotatebox{90}{\large{\textbf{IP \& Copyright}}} \\ \hline

\multirow{9}{*}{\textbf{\large{White-box}}}  
& \large{Model Fine-Tuning} &  &  & \cite{ouyang2022training,rafailov2024direct} & \cite{ouyang2022training,rafailov2024direct} & \cite{ouyang2022training,rafailov2024direct} & \cite{ouyang2022training,rafailov2024direct} & \cite{ouyang2022training,rafailov2024direct} & \cite{ouyang2022training,rafailov2024direct} &  &  &  &  \\ \cline{2-14}
& \large{Architectural Modifications} & \cite{mitchell2021fast,wu2022memorizing,du2022glam,borgeaud2022improving} &  &  & \cite{mitchell2021fast,wu2022memorizing,du2022glam,borgeaud2022improving} & \cite{mitchell2021fast,wu2022memorizing,du2022glam,borgeaud2022improving} & \cite{mitchell2021fast,wu2022memorizing,du2022glam,borgeaud2022improving} & \cite{mitchell2021fast,wu2022memorizing,du2022glam,borgeaud2022improving} &  &  &  &  &  \\ \cline{2-14}
& \large{Bias Mitigation Techniques} &  &  &  &  &  &  &  & \cite{meade2021empirical,zhang2018mitigating} & \cite{meade2021empirical,zhang2018mitigating} &  &  &  \\ \cline{2-14}
& \large{Knowledge Base Integration} &  &  & \cite{borgeaud2022improving} &  & \cite{borgeaud2022improving} & \cite{borgeaud2022improving} & \cite{borgeaud2022improving} &  &  &  &  &  \\ \cline{2-14}
& \large{Regularization Methods} &  &  &  & \cite{sankararaman2022bayesformer} &  &  &  &  &  &  &  &  \\ \cline{2-14}
& \large{Adversarial Training} & \cite{zhang2018mitigating,carlini2019secret} & \cite{zhang2018mitigating,carlini2019secret} &  &  & \cite{zhang2018mitigating,carlini2019secret} &  &  & \cite{zhang2018mitigating,carlini2019secret} &  & \cite{zhang2018mitigating,carlini2019secret} & \cite{zhang2018mitigating,carlini2019secret} &  \\ \cline{2-14}
& \large{Confidence Scoring} &  & \cite{jiang2021can} &  &  &  &  &  &  &  &  &  &  \\ \cline{2-14}
& \large{Formal Verification} & \cite{huang2024survey} &  &  &  &  &  &  &  &  &  &  &  \\ \cline{2-14}
& \large{Customized Loss Functions} &  &  &  & \cite{zhang2018mitigating} & \cite{zhang2018mitigating} &  &  &  &  &  &  &  \\ \hline

\multirow{10}{*}{\textbf{\large{Gray-box}}}  
& \large{Ensemble Modeling} &  & \cite{lakshminarayanan2017simple} &  & \cite{lakshminarayanan2017simple} & \cite{lakshminarayanan2017simple} &  &  &  &  &  &  &  \\ \cline{2-14}
& \large{Human-in-the-Loop Systems} & \cite{christiano2017deep} &  & \cite{christiano2017deep} & \cite{christiano2017deep} & \cite{christiano2017deep} & \cite{christiano2017deep} & \cite{christiano2017deep} & \cite{christiano2017deep} & \cite{christiano2017deep} &  & \cite{christiano2017deep} &  \\ \cline{2-14}
& \large{Retrieval-Augmented Generation} &  &  & \cite{lewis2020retrieval} &  & \cite{lewis2020retrieval} & \cite{lewis2020retrieval} & \cite{lewis2020retrieval} &  &  &  &  &  \\ \cline{2-14}
& \large{Hybrid Feedback Mechanisms} & \cite{perez2022red} & \cite{perez2022red} & \cite{perez2022red} & \cite{perez2022red} & \cite{perez2022red} &  &  &  &  &  &  &  \\ \cline{2-14}
& \large{Semi-Supervised Learning} &  &  & \cite{chen2020big,sohn2020fixmatch} &  & \cite{chen2020big,sohn2020fixmatch} &  &  & \cite{chen2020big,sohn2020fixmatch} &  &  &  &  \\ \cline{2-14}
& \large{External Knowledge Integration} &  &  &  &  & \cite{borgeaud2022improving} & \cite{borgeaud2022improving} & \cite{borgeaud2022improving} &  &  &  &  &  \\ \cline{2-14}
& \large{Dynamic Parameter Adjustment} &  &  &  & \cite{sankararaman2022bayesformer} &  &  &  &  &  &  &  &  \\ \cline{2-14}
& \large{Selective Response Generation} &  & \cite{jiang2021can} &  &  &  &  &  &  &  &  &  &  \\ \cline{2-14}
& \large{Post-Hoc Explainability Tools} & \cite{mitchell2021fast} & \cite{mitchell2021fast} &  &  &  &  &  &  &  &  &  &  \\ \cline{2-14}
& \large{Policy Gradient Fine-Tuning} & \cite{bai2022training} &  &  &  &  &  &  & \cite{bai2022training} & \cite{bai2022training} & \cite{bai2022training} & \cite{bai2022training} &  \\ \hline

\multirow{6}{*}{\textbf{\large{Black-box}}}  
& \large{Output Filtering} &  &  &  & \cite{inan2023llama,guardrails_ai_github} & \cite{inan2023llama,guardrails_ai_github} &  &  &  & \cite{inan2023llama,guardrails_ai_github} &  & \cite{inan2023llama,guardrails_ai_github} &  \\ \cline{2-14}
& \large{Rule-Based Post-Processing} & \cite{rebedea2023nemo} &  & \cite{rebedea2023nemo} & \cite{rebedea2023nemo} & \cite{rebedea2023nemo} & \cite{rebedea2023nemo} & \cite{rebedea2023nemo} & \cite{rebedea2023nemo} & \cite{rebedea2023nemo} & \cite{rebedea2023nemo} & \cite{rebedea2023nemo} & \cite{rebedea2023nemo} \\ \cline{2-14}
& \large{External Fact-Checking} & \cite{manakul2023selfcheckgpt,lewis2020retrieval} &  &  &  & \cite{manakul2023selfcheckgpt,lewis2020retrieval} &  &  &  &  &  &  &  \\ \cline{2-14}
& \large{Content Moderation} &  &  &  &  &  &  &  & \cite{gehman2020realtoxicityprompts,zhang2022opt} & \cite{gehman2020realtoxicityprompts,zhang2022opt} & \cite{gehman2020realtoxicityprompts,zhang2022opt} & \cite{gehman2020realtoxicityprompts,zhang2022opt} &  \\ \cline{2-14}
& \large{Adversarial Input Detection} & \cite{wang2023robustness} & \cite{wang2023robustness} &  &  &  &  &  &  &  & \cite{wang2023robustness} & \cite{wang2023robustness} &  \\ \cline{2-14}
& \large{Plagiarism Detection} &  &  & \cite{carlini2021extracting} &  &  &  &  &  &  &  &  & \cite{carlini2021extracting} \\ \hline

\end{tabular}
\end{adjustbox}
\caption{Implementation Strategies for LLM Guardrails for Science}
\label{table:strategies-dimensions-final}
\end{table}

\textbf{Human-in-the-Loop Systems:} These systems incorporate human oversight \cite{bai2022training} into the LLM’s decision-making process. Human experts can verify outputs in real time, ensuring that the model adheres to scientific standards and ethical guidelines. This approach is particularly effective for managing complex or novel situations where human expertise is essential. \cite{christiano2017deep}

\textbf{Ensemble Modeling:} Ensemble approaches \cite{lakshminarayanan2017simple} combine multiple models or techniques to increase the robustness and reliability of LLM outputs \cite{jiang2023llm}. In the scientific domain, this can be used to cross-check outputs, ensuring greater consistency and accuracy.

\textbf{Hybrid Feedback Mechanisms:} These approaches integrate feedback from both internal model performance and external monitoring systems. For example, a gray-box method could combine formal verification (internal) with rule-based post-processing (external) to ensure that the model's output is not only consistent but also ethical and legally compliant.

\textbf{Retrieval-Augmented Generation:} RAG augments the generative capabilities of LLMs by retrieving relevant information from trusted sources, ensuring that the outputs are factual and reliable. This helps the model produce accurate, context-aware responses in scientific contexts by integrating up-to-date information from peer-reviewed research or scientific databases. \cite{lewis2020retrieval} \cite{edge2024local}

\textbf{Semi-Supervised Learning:} These approaches leverage labeled and unlabeled data, enabling the model to learn from real-world data while still receiving feedback on performance. This hybrid learning method ensures the LLM can continuously improve its outputs over time. \cite{chen2020big} \cite{sohn2020fixmatch}

Gray-box strategies are well-suited for complex scientific domains where models need to be both internally trustworthy and externally validated. By combining internal adjustments with external interventions, these hybrid approaches offer a balanced solution to managing bias, societal risks, and evolving research demands. Techniques like chain-of-thought (CoT) prompting  \cite{wei2022chain}, which enhances the model's reasoning by guiding it to generate more logical, step-by-step responses, improve precision without requiring deep modifications. Additionally, retrieval-augmented generation (RAG) \cite{lewis2020retrieval} allows the model to pull in real-time, trusted external data during the generation process, ensuring outputs are grounded in verified, up-to-date scientific knowledge. These techniques make gray-box methods highly adaptable in handling dynamic scientific challenges. However, they can be resource-intensive, requiring some internal modifications and external interventions. While more flexible than white-box strategies, they may lack the same level of fine-tuned control over the model’s internal processes.

To ensure the effectiveness of these LLM guardrails across diverse applications in the scientific domain, table \ref{table:strategies-dimensions-final} outlines a variety of approaches categorized into white-box, black-box, and gray-box methods. These approaches align specific techniques to key dimensions such as consistency, uncertainty identification, fairness, and attribution, ensuring that each method can address the unique challenges posed by the LLM in scientific contexts. White-box approaches, such as model architecture modification and fine-tuning, enhance internal mechanisms like bias mitigation, formal verification, and uncertainty quantification. These methods are well-suited for tasks requiring high levels of control and trustworthiness, allowing LLMs to handle sensitive scientific data with greater precision.

On the other hand, black-box approaches offer external mechanisms such as rule-based enforcement and API-based moderation, making them ideal for addressing ethical concerns like toxicity, factuality, and societal impact without altering the internal structure of the model. Gray-box methods provide a flexible combination of internal and external techniques, leveraging hybrid strategies such as human-in-the-loop verification and cross-validation with trusted datasets. These approaches, mapped to specific dimensions in the table, provide a balanced solution for managing complex scenarios, such as intellectual property compliance, conflict identification, and time-sensitive outputs. By aligning the appropriate guardrail strategy with the relevant dimension, the framework ensures that LLMs can be governed effectively in the evolving and dynamic landscape of scientific research.

\section{Conclusion}

In this paper, we explored the unique challenges encountered when deploying Large Language Models (LLMs) for scientific domain applications. We found that the sensitive nature of the domain requires specialized guardrails to ensure these LLMs are both trustworthy and useful. We identified critical gaps that the general-purpose guardrails often fail to address, including issues such as hallucinations, time sensitivity, knowledge contextualization, and intellectual property concerns, which can undermine the reliability and integrity of outputs in scientific contexts.

To address these gaps, we propose a comprehensive and systematic framework for guardrailing the LLMs. Our contributions are twofold: First, we developed a structured framework that identifies unique challenges in science, incorporating dimensions such as temporal relevance, contextualized knowledge, and conflict resolution to address the limitations of general-purpose guardrails. Second, we proposed practical implementation strategies that systematically apply these guardrails, leveraging white-box, black-box, and gray-box approaches, supplemented by a review of relevant prior work where applicable.

In future works, we plan to focus on empirically validating these guardrails in real-world scientific applications and refining them to adapt to ongoing advancements in LLMs. This includes testing the effectiveness of our proposed strategies across various downstream scientific disciplines.

\bibliographystyle{IEEEtran}
\bibliography{main}

\end{document}